\documentclass[journal]{IEEEtranTIE}
\usepackage{graphicx}
\usepackage{cite}
\usepackage{picinpar}
\usepackage{amsmath}
\usepackage{url}
\usepackage{flushend}
\usepackage[latin1]{inputenc}
\usepackage{colortbl}
\usepackage{soul}
\usepackage{multirow}
\usepackage{pifont}
\usepackage{color}
\usepackage{alltt}
\usepackage[hidelinks]{hyperref}
\usepackage{enumerate}
\usepackage{siunitx}
\usepackage{breakurl}
\usepackage{epstopdf}
\usepackage{pbox}
\usepackage{amsmath} 
\usepackage{amssymb}  

\begin{document}
\title{	Load-Aware Locomotion Control for Humanoid Robots in Industrial Transportation Tasks}

\author{
	\vskip 1em
	Lequn Fu,
	Yijun Zhong,
        Xiao Li,
        Yibin Liu,
        Zhiyuan Xu,
        Jian Tang,~\emph{Fellow, IEEE},
	and Shiqi Li
	\thanks{
	
		
        Lequn Fu, Yijun Zhong, Xiao Li, and Yibin Liu are with the Huazhong University of Science and Technology, Wuhan 430074, China 
        (e-mail: fulq@hust.edu.cn; zhongyijun@hust.edu.cn; li\_xiao@hust.edu.cn; liu\_yibin@hust.edu.cn).
        
        Zhiyuan Xu, and Jian Tang are with the Beijing Humanoid Robot Innovation Center, Beijing 100086, China 
        (e-mail: eric.xu@x-humanoid.com; jian.tang@x-humanoid.com).

		Shiqi Li is with the Huazhong University of Science and Technology, Wuhan 430074, China
        (e-mail: sqli@hust.edu.cn).
	}
}

\maketitle

\begin{abstract}
Humanoid robots deployed in industrial environments are required to perform load-carrying transportation tasks that tightly couple locomotion and manipulation. However, achieving stable and robust locomotion under varying payloads and upper-body motions is challenging due to dynamic coupling and partial observability. This paper presents a load-aware locomotion framework for industrial humanoids based on a decoupled yet coordinated loco-manipulation architecture. Lower-body locomotion is controlled via a reinforcement learning policy producing residual joint actions on kinematically derived nominal configurations. A kinematics-based locomotion reference with a height-conditioned joint-space offset guides learning, while a history-based state estimator infers base linear velocity and height and encodes residual load- and manipulation-induced disturbances in a compact latent representation. The framework is trained entirely in simulation and deployed on a full-size humanoid robot without fine-tuning. Simulation and real-world experiments demonstrate faster training, accurate height tracking, and stable loco-manipulation. Project page: \url{https://lequn-f.github.io/LALO/}.
\end{abstract}


\markboth{IEEE TRANSACTIONS ON INDUSTRIAL ELECTRONICS}%
{}

\definecolor{limegreen}{rgb}{0.2, 0.8, 0.2}
\definecolor{forestgreen}{rgb}{0.13, 0.55, 0.13}
\definecolor{greenhtml}{rgb}{0.0, 0.5, 0.0}

\section{Introduction}
\IEEEPARstart{H}{umanoid} robots are increasingly expected to operate in industrial environments, where they must perform load-carrying transportation tasks such as box handling, material transfer, and workstation-to-workstation delivery~\cite{Gu2025humanoid,Sheng2025comprehensivea}.
Unlike locomotion-only scenarios, industrial box-handling tasks require tight coordination between locomotion and upper-body manipulation while maintaining balance and safety under varying payloads.
During squatting, lifting, carrying, and walking motions, the payload mass, location, and distribution can change significantly, introducing strong dynamic coupling between the upper and lower body.
These variations induce time-varying disturbances, shifts in the center of mass, and changes in ground reaction forces, posing substantial challenges to stable humanoid locomotion.

From a control perspective, load-carrying locomotion has traditionally been addressed using model-based whole-body control and optimization-based frameworks~\cite{Rigo2024hierarchical,Qi2025a,He2024cdmmpc,Yamamoto2020survey,Winkler2018gait}.
By explicitly modeling payload effects as changes in mass distribution or external wrenches, such methods can, in principle, account for manipulation-induced dynamics~\cite{Chignoli2021mit,Calvert2026behavior}.
However, their performance is highly sensitive to modeling accuracy, contact estimation, and payload identification, and often requires careful tuning for specific task configurations.
In realistic industrial settings, payload properties may be unknown, vary across tasks, or change during operation, making it difficult for purely model-based approaches to maintain robust performance.

Recent advances in reinforcement learning (RL) have demonstrated strong potential for humanoid locomotion by directly optimizing feedback policies in high-dimensional state spaces~\cite{Gangapurwala2022rloc,Kim2024not,Zhang2025chaos,Bao2025deep,Jenelten2024dtc,Li2025reinforcement,Marum2024revisiting,Zhou2026curriculum}. Learning-based controllers are inherently more tolerant to modeling errors and contact uncertainties.

Research on humanoid loco-manipulation has generally followed two primary paradigms. The first paradigm adopts a decoupled control strategy, where reinforcement learning is used to generate lower-body locomotion while upper-body motions are produced by kinematic or optimization-based controllers~\cite{Ji2026a,Li2025thor,Cheng2024expressive,Lu2025mobile}. 
For example, Fu et al.~\cite{Fu2022deep} proposed a unified learning framework for legged manipulators but treated arm and leg control separately, which limited the consideration of manipulation-induced disturbances on locomotion stability.
Murooka et al.~\cite{Murooka2025tact} integrated imitation learning with tactile feedback for whole-body contact manipulation; however, locomotion remained model-based and retargeted, and the dynamic influence of arm motion on bipedal balance was not explicitly addressed.
Xue et al.~\cite{Xue2025unified} presented a versatile whole-body controller supporting external upper-body interventions such as teleoperation, yet its decoupled formulation does not allow the lower-body policy to adapt online to payload variations induced by arm motion.
More broadly, many decoupled approaches simplify payload effects by modeling the load as a static mass attached to the torso~\cite{Fu2026humaninspired}, overlooking a key characteristic of industrial box-handling tasks: payloads are carried by upper-body end-effectors and move relative to the base as the arms and torso articulate~\cite{Chappellet2024humanoid}.
Moreover, base-height regulation in these methods is often realized through hard tracking or direct constraint enforcement, which lacks flexibility under varying payloads and may degrade contact smoothness during squatting and lifting.
Although some recent works incorporate upper-body observations into locomotion policies~\cite{Ben2025homie,Zhang2025falcon}, they often lack precise feedback mechanisms for autonomous squatting and lifting height regulation, and frequently rely on teleoperation rather than fully autonomous loco-manipulation.

The second paradigm directly trains whole-body motions using reinforcement learning, aiming to implicitly capture inter-limb coordination and contact-rich behaviors~\cite{Fu2024humanplus,He2024omnih2o,He2024learning,Ji2025exbody2,Elobaid2023online,Kuang2025skillblender,Tessler2025maskedmanipulator}.
Zhang et al.~\cite{Zhang2024wococo} proposed the WoCoCo framework, which decomposes tasks into sequential contact stages and demonstrates impressive versatility across box loco-manipulation, dynamic dancing, and cliff climbing.
However, the framework does not explicitly address payload variability or provide dedicated mechanisms for posture and height regulation under changing load conditions. Similarly, Dao et al.~\cite{Dao2024simtoreal} demonstrated sim-to-real transfer of box loco-manipulation on the Digit humanoid by decomposing the task into multiple modular policies, but relied on predefined contact sequences and exhibited limited adaptability to unmodeled payload variations.
Despite their expressiveness, existing whole-body RL approaches rarely model the dynamic interplay between upper-body payload motion and lower-body locomotion in an explicit or structured manner, which restricts robustness in realistic industrial scenarios.

In contrast to existing approaches, this work addresses both dynamic upper--lower body coupling and partial observability within a unified locomotion framework. 
We propose a decoupled yet tightly coordinated locomotion--manipulation architecture, where reinforcement learning is applied exclusively to lower-body locomotion while upper-body box-handling motions are generated by a perception-driven kinematic controller and incorporated into the locomotion policy observations. 
The locomotion controller is further guided by a kinematics-based reference conditioned on commanded velocity and base height, providing structured motion priors that facilitate stable locomotion under varying payloads. 
To cope with partial observability, a history-based state estimator reconstructs base velocity and height while learning a compact latent representation of load- and manipulation-induced disturbances. 
The framework is trained entirely in simulation with domain randomization and deployed on a full-size humanoid robot without additional fine-tuning, demonstrating robust locomotion in real-world industrial box-handling tasks.

The main contributions of this work are summarized as follows:
\begin{enumerate}
    \item A load-aware locomotion control framework for humanoid box-handling tasks that decouples lower-body locomotion learning from upper-body manipulation while maintaining dynamic coupling through observation design.
    \item A kinematics-based locomotion reference and height-conditioned joint-space offset that enable structured residual reinforcement learning, improving training efficiency, contact quality, and height regulation performance.
    \item A history-based state estimation scheme that explicitly infers base linear velocity and base height while learning a compact latent representation to capture load- and manipulation-induced disturbances under partial observability.
\end{enumerate}

\section{Problem Formulation}
\subsection{Industrial Box-Handling Task}

Humanoid robots operating in industrial logistics environments must perform continuous box-handling cycles, which may include approaching a palletized or non-palletized box, executing a depalletizing motion when required that involves coordinated squatting and reaching, transporting the grasped box under a varying payload, placing it at a designated area, and returning for the next item. These operations introduce substantial variations in upper-body posture and external loading conditions, requiring the robot to maintain dynamic stability throughout.

The robot is modeled as a floating-base articulated system with generalized coordinates $q=[q_b, q_{\text{leg}}, q_{\text{arm}}] \in \mathbb{R}^{n}$, where $q_b$ denotes the 6-DoF floating base, $q_{\text{leg}}$ corresponds to the 12 lower-limb joints, and $q_{\text{arm}}$ to the 14 upper-body joints. When a box is grasped, the payload becomes rigidly coupled to the torso, resulting in unknown variations in mass distribution and center-of-mass (CoM) location. The resulting dynamics are described by $M(q,\Delta m)\ddot{q} + h(q,\dot{q},\Delta m) = S^T \tau + J_c^T \lambda$, with $\Delta m$ denoting the unknown payload parameters. These load-induced disturbances strongly affect torso stabilization and feasible foot placement, especially during squatting and stepping. In the system architecture, only the 12 lower-limb joints are driven by the locomotion controller, whereas the 14 upper-body joints follow trajectories produced by an IK-based manipulation module using target poses $T_{\text{box}} \in SE(3)$. The induced arm motions and payload shifts must therefore be compensated implicitly via proprioceptive and IMU feedback.

\subsection{Reinforcement Learning Formulation}

The locomotion controller is formulated as a reinforcement learning problem defined by the tuple $\mathcal{M} = \langle \mathcal{S}, \mathcal{A}, \mathcal{O}, T, R, \gamma \rangle$. 
Due to sensing limitations and the presence of unknown payloads and manipulation-induced disturbances, the locomotion control problem is inherently partially observable. 
The agent therefore receives only a partial observation $o_t = \psi(s_t)$, which contains proprioceptive measurements such as IMU orientation, angular velocity, joint positions and velocities and locomotion commands.
Table~\ref{tab:observation} summarizes the observation space used by the locomotion policy.

\begin{table}[t]
\centering
\caption{Observation space of the locomotion policy.}
\label{tab:observation}
\begin{tabular}{lcc}
\hline\hline
\textbf{Category} & \textbf{Dimension} & \textbf{Noise Scale} \\ 
\hline
Phase Input & 2 & 0 \\ 
Projected gravity & 3 & $\pm 0.05$ \\ 
Angular velocity & 3 & $\pm 0.1$ \\ 
Joint positions & 27 & $\pm 0.02$ \\ 
Joint velocities & 27 & $\pm 0.5$ \\ 
Command & 4 & 0 \\ 
Last actions & 12 & 0 \\ 
Estimated base velocity & 3 & 0 \\ 
Estimated base height & 1 & 0 \\ 
Latent feature & 32 & 0 \\ 
\hline\hline
\end{tabular}
\end{table}

The control action $a_t \in \mathbb{R}^{12}$ corresponds to the desired joint targets for the 12 lower-limb actuators. 
Because the transition dynamics depend on both the unmodeled payload $\Delta m$ and the time-varying manipulation motions, the locomotion subsystem constitutes a Partially Observable Markov Decision Process (POMDP). 
The objective is to learn a policy $\pi(a_t \mid o_{\le t})$ that maximizes the expected discounted return
\begin{equation}
J(\pi) = \mathbb{E}\left[ \sum_{t=0}^{T} \gamma^{t} r_t \right],
\end{equation}
where the reward function is designed to encourage stable walking, robust balance during squatting and stepping, energy efficiency, and resilience to disturbances arising from box manipulation and payload shifts.

\section{Methods}
\label{sec:methods}

\subsection{Overview}
\label{subsec:overview}

\begin{figure*}[t]
    \vspace{2mm}
    \centering
    \includegraphics[width=1\linewidth]
    {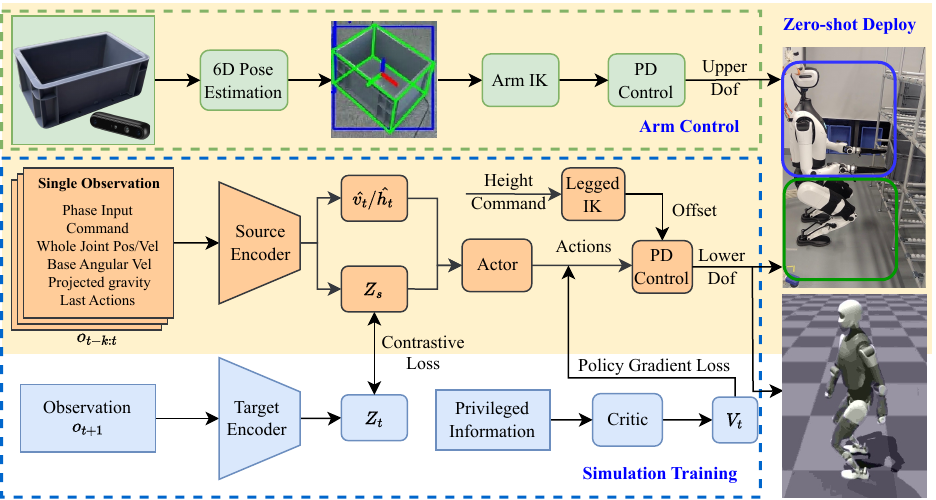}
    \caption{Overview of the proposed load-aware humanoid loco-manipulation framework. Upper-body manipulation is generated by perception-driven kinematic control using 6D object pose estimation, while lower-body locomotion is governed by a residual reinforcement learning policy with height-conditioned offsets and history-based state estimation. Historical proprioceptive observations are encoded by a history-based state estimator to infer base velocity, base height, and a compact latent state capturing load- and manipulation-induced disturbances.}
    \label{fig:Framework}
    \vspace{-2mm}
\end{figure*}
As illustrated in Fig.~\ref{fig:Framework}, we propose a load-aware locomotion framework for humanoid robots performing industrial transportation tasks such as box handling and load carrying. 
The system adopts a decoupled yet tightly coordinated locomotion--manipulation architecture, where lower-body locomotion is controlled by a reinforcement learning policy while upper-body manipulation is executed through perception-driven kinematic control.

At each control step $t$, the robot receives a high-level locomotion command
\begin{equation}
c_t = [v_x^\ast, v_y^\ast, \omega_z^\ast, z_b^\ast],
\end{equation}
where $(v_x^\ast, v_y^\ast)$ and $\omega_z^\ast$ denote the desired planar linear velocity and yaw rate, and $z_b^\ast$ specifies the commanded base height. 
The locomotion policy outputs joint-space position targets for the lower-body joints, which are tracked by joint-level PD controllers. 
Upper-body motions are generated independently using real-time object pose estimation and inverse kinematics.

Although the locomotion policy does not directly command arm joints, upper-body states are included in the policy observation, allowing the controller to react to manipulation-induced disturbances and payload-dependent dynamics.

The framework consists of four main components: 
(i) a kinematics-based locomotion reference generator, 
(ii) a history-based state estimator for partial observability, 
(iii) a reinforcement learning locomotion policy with structured reward design, and 
(iv) an upper-body curriculum strategy that gradually increases dynamic coupling during training.

\subsection{Height-Conditioned Offset Control with Kinematic Reference}
\label{subsec:kin_ref}
To improve training stability, base height tracking accuracy, and contact smoothness, we adopt a height-conditioned offset control formulation augmented with a kinematics-based locomotion reference. In this design, the joint-space offset is determined solely by the commanded base height, while the kinematic reference trajectory is generated from both velocity and height commands and is used only as a training regularizer.

Specifically, at each control step, the commanded base height $z_b^\ast$ is mapped to a nominal lower-body joint configuration through inverse kinematics. This configuration defines a height-conditioned joint offset
\begin{equation}
q_{leg}^{off}(t) = \mathrm{IK}\!\left(z_b^\ast\right),
\end{equation}
which varies continuously with the height command and provides a physically consistent vertical posture for the robot.
The offset does not encode stepping or gait patterns and is independent of velocity commands.

On top of this offset, the locomotion policy outputs joint-space residual actions $\Delta q_{leg}(t)$, which are added to form the final joint position commands tracked by low-level PD controllers,
\begin{equation}
q_{leg}^{cmd}(t)=q_{leg}^{off}(t)+\Delta q_{leg}(t).
\end{equation}
This formulation allows the policy to focus on local corrections around a height-consistent nominal posture, rather than learning absolute joint targets across varying height conditions.

In parallel, a kinematics-based locomotion reference trajectory is generated from the commanded planar velocity $(v_x^\ast, v_y^\ast)$, yaw rate $\omega_z^\ast$, and base height $z_b^\ast$. A normalized gait phase is defined as
\begin{equation}
\varphi_t = \frac{t \Delta t}{T_c} + \varphi_0,
\end{equation}
where $T_c$ denotes the gait cycle duration and $\varphi_0$ is a random phase offset.
Based on the gait phase and velocity commands, foot-end trajectories are generated for each leg. During the swing phase, the vertical foot motion is parameterized by a smooth cubic polynomial over normalized swing time $\tau \in [0,1]$,
\begin{equation}
z_{sw}^{ref}(\tau)=a_0+a_1\tau+a_2\tau^2+a_3\tau^3,
\end{equation}
which satisfies zero position and velocity boundary conditions at lift-off and touch-down. The peak swing clearance is modulated by a task-dependent height parameter, ensuring smooth ground contact and reduced impact forces. During the stance phase, the foot height remains consistent with the commanded base height $z_b^\ast$.

The resulting reference joint trajectories $q_{leg}^{ref}(t)$, obtained via inverse kinematics from the foot-end trajectories,
are \emph{not} tracked by the controller. Instead, they are incorporated only through auxiliary reward terms
on joint motion and foot-end kinematics during training. At deployment, the controller relies solely on the height-conditioned offset and the learned residual policy.
By separating height regulation (offset), locomotion structure (reference), and disturbance compensation (residual policy),
this formulation improves convergence speed, enhances base height tracking under varying commands, and leads to smoother foot--ground contacts without restricting policy expressiveness.

\subsection{History-Based State Estimation}
\label{subsec:state_est}

Industrial transportation tasks exhibit partial observability due to unmodeled payload dynamics and manipulation-induced disturbances. 
Because such effects cannot be reliably inferred from single-step proprioceptive measurements, we introduce a history-based state estimator that leverages short-term temporal context.

At each control step, a fixed-length window of raw observations is stacked as
\begin{equation}
O_t = [o_{t-T+1}^{raw}, \dots, o_t^{raw}],
\end{equation}
and processed by a lightweight neural encoder. 
The estimator predicts the base linear velocity $\hat{v}_{b,t}$, the base height $\hat{z}_{b,t}$, and a compact latent representation $z_t$ that captures residual dynamic variations caused by payload changes and upper-body motion. 
The latent feature $z_t\in\mathbb{R}^D$ is $\ell_2$-normalized to ensure bounded representations,
\begin{equation}
z_t \leftarrow \frac{z_t}{\|z_t\|_2}.
\end{equation}

The inferred states are incorporated into the policy observation as $[\hat{v}_{b,t}, \hat{z}_{b,t}, z_t]$, enabling the locomotion controller to reason about unobserved dynamics under varying load conditions.

During training, the estimator is supervised using privileged simulation information through regression losses on base linear velocity and base height,
\begin{equation}
\mathcal{L}_{\mathrm{vel}} = \| \hat{v}_{b,t} - v_{b,t} \|_2^2, \quad
\mathcal{L}_{\mathrm{height}} = \| \hat{z}_{b,t} - z_{b,t} \|_2^2.
\end{equation}

where $v_{b,t}$ and $z_{b,t}$ denote the ground-truth base linear velocity and base height, respectively. 
To encourage informative latent representations, we additionally employ a prototype-based swap prediction objective with Sinkhorn-normalized assignments.
The overall estimator objective is
\begin{equation}
\mathcal{L}_{\mathrm{est}} =
\mathcal{L}_{\mathrm{vel}} +
\mathcal{L}_{\mathrm{height}} +
\mathcal{L}_{\mathrm{aux}}.
\end{equation}

During deployment, the estimator operates solely on onboard sensory inputs and serves as a preprocessing module for the locomotion policy.

\subsection{End-to-End Deep Reinforcement Learning}
\label{subsec:rl}

\begin{table}[t]
\centering
\caption{Reward terms and weights used for training.}
\label{tab:reward_term}
\small
\renewcommand{\arraystretch}{1.3}
\resizebox{\columnwidth}{!}{%
\begin{tabular}{l l c}
\hline\hline \\[-3mm]
\textbf{Reward term} & \textbf{Formulation} & \textbf{Weight} $(\mu_i)$ \\
\hline
Linear velocity tracking &
$\exp\!\left(-4\|[v_x,v_y]-[v_x^\ast,v_y^\ast]\|^2\right)$ &
$1.2$ \\

Angular velocity tracking &
$\exp\!\left(-4(\omega_z-\omega_z^\ast)^2\right)$ &
$1.1$ \\

Base height tracking &
$\exp(-50\,|h_{b,rel}-h_{b,rel}^\ast|)$ &
$2.0$ \\

Low-speed consistency &
$r_{\mathrm{ls}}(v_x,v_x^\ast)\cdot \mathbb{I}(|v_x^\ast|>0.1)$ &
$0.2$ \\

Orientation stabilization &
$\exp\!\left(-10\,\|g_{proj,xy}\|\right)$ &
$1.0$ \\

Velocity mismatch &
$0.5(\exp(-10 v_z^2)+\exp(-5\|\omega_{x,y}\|))$ &
$0.5$ \\

Base acceleration &
$\exp(-\|\Delta v_{\mathrm{root}}\|)$ &
$0.2$ \\

Feet contact consistency &
$0.5\!\sum_{f\in\{L,R\}}(\mathbb{I}[c_f=s_f]-0.5\,\mathbb{I}[c_f\neq s_f])$ &
$1.2$ \\

Joint imitation &
$\exp(-2\|q_{leg}-q_{leg}^{ref}\|)$ &
$1.0$ \\

Foot height tracking &
$\sum_{f\in\{L,R\}}\mathbb{I}(h^{ref}_{f}-h_{f}<0.008)$ &
$1.0$ \\

Foot slip &
$\sum_{f\in\{L,R\}}\|\mathbf{v}_{foot,f}^{xy}\|\cdot \mathbb{I}[F_{z,f}>1]$ &
$-0.25$ \\

Foot stumble &
$\mathbb{I}\!\left(\exists f:\ \|F_{xy,f}\|>3|F_{z,f}|\right)$ &
$-1.5$ \\

Contact force &
$\sum_{f\in\{L,R\}}\big(\|F_f\|-F_{\max}\big)_+$ &
$-0.005$ \\

Contact momentum &
$\sum_{f\in\{L,R\}}\min(v_{z,f},0)\cdot \max(F_{z,f}-50,0)$ &
$-0.005$ \\

Hip-joint deviation &
$\exp\!\left(-50\,\delta_{hip}\right)-0.01\,\|q_{leg}-q_{leg}^0\|$ &
$0.5$ \\

Action smoothness &
$\|a_t-a_{t-1}\|^2 + \|a_t+a_{t-2}-2a_{t-1}\|^2$ &
$-0.02$ \\

Torque penalty &
$\|\tau\|^2$ &
$-10^{-5}$ \\

Joint velocity &
$\|\dot q\|^2$ &
$-10^{-4}$ \\

Joint acceleration &
$\left\|(\dot q_t-\dot q_{t-1})/\Delta t\right\|^2$ &
$-10^{-7}$ \\

Velocity limit penalty &
$\sum_j\left(|\dot q_j|-\kappa_v\,\dot q^{\max}_j\right)_+$ &
$-0.1$ \\

Torque limit penalty &
$\sum_j\left(|\tau_j|-\kappa_\tau\,\tau^{\max}_j\right)_+$ &
$-0.1$ \\

Position limit penalty &
$\sum_j\left((q_j-q_j^{\max})_+ + (q_j^{\min}-q_j)_+\right)$ &
$-2.0$ \\

Stand-still &
$\sum_j\left|q_j-q^{\text{stand}}_j\right|
+\mathbb{I}\left(|z-h^{cmd}|<0.01\right)$ &
$-3.0$ \\

Collision penalty &
$\sum \mathbb{I}[\mathrm{undesired\ contact}]$ &
$-1$ \\
\hline\hline
\end{tabular}}
\end{table}

The lower-body locomotion controller is trained in an end-to-end manner. At each control step, the policy outputs joint-space action commands as position offsets with respect to a nominal configuration.

\subsubsection{Reward Design}
\label{subsubsec:reward}

The reward function is designed to balance command tracking accuracy, dynamic stability, contact quality, and motion regularity. At time step $t$, the total reward is defined as
\begin{equation}
r_t = \sum_i \mu_i r_i,
\end{equation}
where $r_i$ denotes an individual reward term and $\mu_i$ is its corresponding weight.
we employ exponentially shaped rewards to emphasize small residual errors while maintaining smooth and well-conditioned gradients. The shaping coefficients are selected individually according to the sensitivity of each task.

Table~\ref{tab:reward_term} summarizes all reward terms used for training. In the table, $v_x$ and $v_y$ denote the planar base linear velocities, $\omega_z$ denotes the yaw angular velocity, and $(\cdot)^\ast$ represents commanded values. The base height $h_{b,rel}$ is measured relative to the supporting feet.
Kinematics-based reference terms, including joint imitation and foot height tracking,
are incorporated only during training as soft constraints to guide exploration and improve convergence,
while remaining inactive during deployment. Contact-related penalties suppress excessive impact forces,
foot slip, unstable contacts, and undesired collisions, thereby improving contact smoothness and robustness.
Additional regularization terms on action variation, joint motion, and actuation effort are included to promote smooth and energy-efficient behaviors.

\subsubsection{Upper-Body Curriculum}
\label{subsec:curriculum}

Although the locomotion policy directly controls only the lower-body joints, upper-body motions can significantly affect whole-body dynamics through inertial coupling, especially during load-carrying tasks.
Injecting strong upper-body disturbances from the early training stage often leads to unstable learning and degraded convergence. To mitigate this issue, we adopt an upper-body curriculum strategy.

During training, random upper-body joint motions are applied as external disturbances with bounded magnitude
and smooth temporal interpolation. The disturbance intensity is governed by a scalar ratio $\rho\in[0,1]$,
which is progressively increased as the policy achieves stable locomotion performance. This curriculum sampling scheme follows the strategy introduced in~\cite{Ben2025homie}, allowing the policy to first acquire robust lower-body locomotion and then gradually adapt to stronger whole-body perturbations.
During deployment, the curriculum randomization is disabled.

\subsubsection{Domain Randomization}
To reduce the sim-to-real gap, we apply domain randomization during training so that the locomotion policy is exposed to a broad range of uncertainties that commonly arise in real deployment. The randomized factors include variations in contact and environment properties, robot inertial parameters, actuation imperfections, and control timing. In addition, intermittent external pushes are applied to improve disturbance rejection. This combination encourages the policy to rely on robust feedback strategies rather than exploiting simulator-specific dynamics.
Concretely, we randomize ground friction and restitution to cover diverse contact conditions. To model variations caused by loads and modeling errors, we perturb payload masses and apply center-of-mass (CoM) offsets, together with link-mass scaling. To account for hardware inconsistencies, we randomize PD gains and inject additive actuation offsets and torque perturbations. Finally, we introduce stochastic action execution delay to emulate control-level latency. The domain randomization settings are summarized in Table~\ref{tab:domain_rand}.

\begin{table}[!t]
\renewcommand{\arraystretch}{1.3}
\caption{Overview of Domain Randomization}
\label{tab:domain_rand}
\centering
\begin{tabular}{lccc}
\hline\hline \\[-3mm]
\textbf{Parameter} & \textbf{Unit} & \textbf{Range} & \textbf{Operator} \\
\hline 
Ground Friction & -- & [0.1, 3.0] & -- \\
Restitution Coefficient & -- & [0.0, 1] & -- \\
External Push Velocity (XY) & m/s & [-0.5, 0.5] & additive \\
External Push Angular Velocity & rad/s & [0, 0.6] & additive \\
Base Mass Offset & kg & [-5.0, 10.0] & additive \\
Hand Mass Offset & kg & [-0.1, 0.3] & additive \\
Base CoM Offset & m & [-0.03, 0.03] & additive \\
Link Mass Scaling & -- & [0.8, 1.0] & scaling \\
Joint Position Offset & rad & [-0.05, 0.05] & additive \\
Joint Friction & -- & [0.01, 1.15] & -- \\
Joint Armature & -- & [0.008, 0.06] & -- \\
PD Factors & -- & [0.9, 1.1] & scaling \\
Torque Output Scaling & -- & [0.8, 1.2] & scaling \\
Action Latency & ms & [10, 50] & -- \\
Motor/IMU Observation Latency & ms & [10, 50] & -- \\
\hline\hline
\end{tabular}
\end{table}

\subsection{Upper-Body Control and Deployment}
\label{subsec:upper_body}

\begin{figure}[t]
    \centering
    \includegraphics[width=\linewidth]{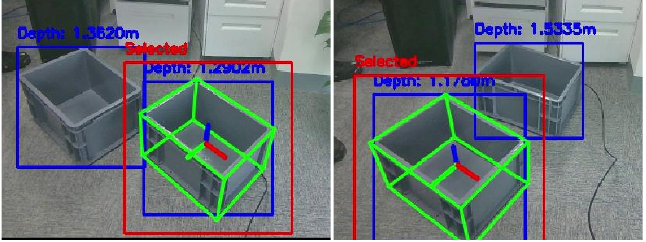}
    \caption{Multi-box detection and target 6D pose estimation. 
    Blue boxes denote instances detected by YOLO-seg, the red box indicates the selected target, and the green wireframe shows the 6D pose estimated by FoundationPose.}
    \label{fig:multi_box_pose}
\end{figure}

In real-world deployment, upper-body manipulation is executed independently of the locomotion policy through a perception-driven kinematic control pipeline.
To handle scenarios with multiple identical boxes in the workspace, we integrate instance segmentation with a 6D pose estimation framework to enable multi-object detection and target selection.
As shown in Fig.~\ref{fig:multi_box_pose}, a YOLO-based~\cite{Wang2024yolov10} segmentation network is first applied to the RGB image to obtain pixel-wise masks of all visible boxes. According to task requirements, a target box is selected from the segmented instances (e.g., the nearest box or the leftmost/rightmost instance).
The corresponding bounding region in the RGB image is then cropped and fed into a FoundationPose-based~\cite{Wen2024foundationpose} 6D pose estimator, which outputs the object pose relative to the camera frame.
When object configurations change due to environmental interactions, the detection and pose estimates are updated in real time to maintain tracking consistency.
Given the estimated 6D pose, a predefined grasp or pre-grasp pose is generated, and inverse kinematics is used to compute the corresponding arm joint trajectories.
The resulting joint commands are executed by position-based controllers to accomplish grasping, lifting, and placement motions.

This loop runs concurrently with the learned locomotion controller. By decoupling upper-body planning from lower-body learning, the system ensures precise and predictable manipulation behavior, while the locomotion policy adapts online to manipulation-induced dynamic effects through its observation inputs, enabling stable load-carrying locomotion in industrial environments.

\section{Experiments}
\label{sec:experiments}

\subsection{System Overview and Implementation Details}
\label{subsec:system_impl}

All simulation and real-world experiments are conducted on Tiangong~2.0 Pro, a full-size humanoid robot developed by the Beijing Humanoid Robot Innovation Center.
The robot is $1.73$~m tall, weighs $74$~kg, and is fully electrically actuated with $42$ degrees of freedom (DoFs), including legs, arms, waist, head, and dexterous hands.

For locomotion and loco-manipulation experiments, a subset of $27$ DoFs is activated, consisting of $6$ DoFs per leg, $7$ upper-body DoFs, and $1$ waist DoF.
Head joints are fixed, and the original hands are replaced with a custom end-effector designed for robust box handling.
All active joints are controlled by position-based PD controllers at $500$~Hz, while the reinforcement learning policy runs at $100$~Hz.
Upper-body manipulation is executed independently using a perception-driven kinematic controller.
Object perception is provided by a head-mounted depth camera for box localization and pose estimation.
All policies are trained in simulation using Proximal Policy Optimization (PPO)~\cite{Schulman2017} with extensive domain randomization and are deployed on the real robot without additional fine-tuning.
Training is performed in Isaac Gym with 4096 parallel environments on a single NVIDIA RTX~4090 GPU.
The hyperparameters are listed in Table~\ref{tab:hyperparameters}.

\begin{table}[htbp]
\centering
\vspace{-2mm}
\caption{PPO Training Hyperparameters}
\label{tab:hyperparameters}
\begin{tabular}{ll}
\hline\hline 
Parameter Name & Value \\
\hline
Value loss coefficient & 1.0 \\
Number of Epochs & 5 \\
Clipping & 0.2 \\
Entropy Coefficient & 0.003 \\
Discount Factor & 0.99 \\
GAE Discount Factor & 0.95 \\
Desired KL-Divergence & 0.01 \\
Learning Rate & 1e-3 (adaptive) \\
Actor hidden layers & [512, 256, 256]\\
Source Encoder hidden layers & [256, 256]\\
Target Encoder hidden layers & [256, 256]\\
Number of History Observation & 6 \\
\hline\hline
\end{tabular}
\end{table}

\subsection{Effect of Kinematic Reference with Residual Control}
\label{subsec:residual_ref}

To evaluate the effectiveness of the proposed height-conditioned offset and kinematics-based locomotion reference, we conduct an ablation study comparing three locomotion controllers. All methods share the same policy architecture, training procedure, and domain randomization settings, and differ only in the formulation of the joint offset and the use of kinematics-based reference trajectories.
The reward structure is kept consistent across methods, while the weight of the base-height tracking term is adjusted when necessary to ensure that each controller can successfully follow the commanded height, such that all methods are evaluated on the same task. For each method, policies are trained with three different random seeds. Evaluation is carried out in 2000 parallel environments, each running for 20~s under randomly sampled velocity and base height commands. Reported results are obtained by averaging across the three trained policies, with the standard deviation indicating performance variation.

Specifically, we consider three methods.
\paragraph{Fixed Offset w/o Reference} serves as the baseline, where a fixed nominal joint configuration is used as the offset
and no kinematics-based reference is provided, resulting in a purely learning-based controller with unstructured exploration.
\paragraph{Ours Fixed Offset} augments the baseline with kinematics-based reference trajectories generated from the commanded velocity and base height, which are incorporated only through auxiliary reward terms during training, while keeping the joint offset fixed.
\paragraph{Ours} further replaces the fixed offset with a height-conditioned joint-space offset computed via inverse kinematics from the commanded base height, on top of which the policy outputs residual joint actions; the same kinematics-based reference rewards are retained.

\begin{figure}[t]
    \vspace{2mm}
    \centering
    \includegraphics[width=1\linewidth]
    {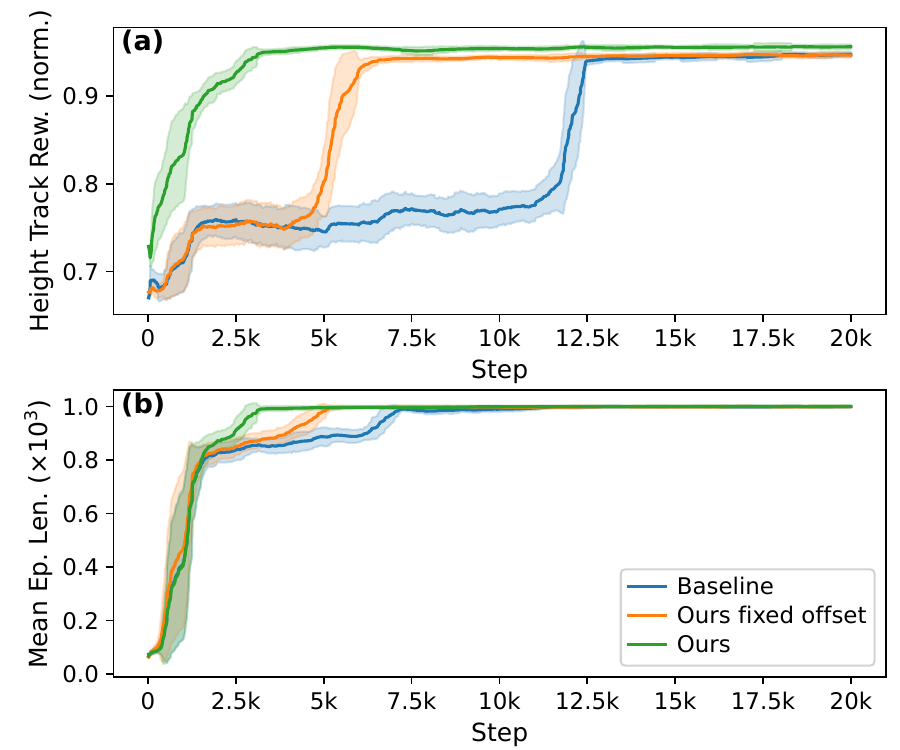}
    \caption{Training convergence comparison of different offset and reference formulations. (a) Episode reward for tracking base height versus environment steps. (b) Mean episode length versus environment steps.}
    \label{fig: ablation_convergence}
    \vspace{-2mm}
\end{figure}
As shown in Fig.~\ref{fig: ablation_convergence}, incorporating kinematics priors consistently improves training efficiency and stability compared to the purely learning-based baseline (Fixed Offset w/o Reference).
In Fig.~\ref{fig: ablation_convergence}(a), the baseline converges slowly with large reward fluctuations, indicating unstructured exploration and difficulty in satisfying the base-height tracking objective.
Introducing kinematics-based reference rewards accelerates reward growth and reduces variance, demonstrating their effectiveness in guiding early-stage learning.
Replacing the fixed nominal offset with a height-conditioned joint-space offset further improves training dynamics, leading to the fastest and most stable convergence.
By providing a physically plausible nominal configuration, this formulation allows the policy to focus on learning residual joint corrections rather than discovering postures from scratch, thereby improving sample efficiency and robustness.
A consistent trend is observed in Fig.~\ref{fig: ablation_convergence}(b).
Both the reference-based and proposed methods achieve longer and more stable episode lengths than the baseline, indicating fewer premature terminations during training.
The proposed method attains the longest and most consistent episodes, confirming that reference shaping and height-conditioned offsets provide complementary benefits for robust locomotion learning.

\begin{figure}[t]
    \vspace{2mm}
    \centering
    \includegraphics[width=1\linewidth]
    {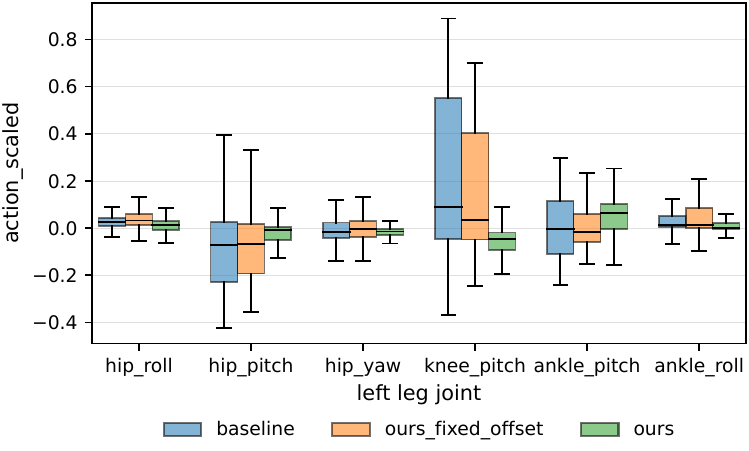}
    \caption{Distribution of joint residual actions during steady-state walking.}
    \label{fig:action_distribution}
    \vspace{-2mm}
\end{figure}
Fig.~\ref{fig:action_distribution} compares the action distributions of left-leg joints under different offset and reference formulations. The baseline controller exhibits wide and irregular action distributions across multiple joints, reflecting unstructured exploration and a strong reliance on policy outputs
to compensate for posture and height variations. Introducing kinematics-based reference rewards slightly regularizes the action distributions, but the overall spread remains comparable, indicating that reference shaping alone does not fundamentally change the underlying action parameterization.

In contrast, the proposed method produces more structured and joint-dependent action distributions. Notably, actions concentrate around joint-specific operating regions with reduced bias and improved symmetry across the kinematic chain. This behavior suggests that the height-conditioned inverse-kinematics offset absorbs most posture- and height-related variations, allowing the policy to focus on learning residual corrections rather than compensating for global configuration changes. As a result, the learned actions exhibit improved regularity and coordination, which is consistent with the observed gains in training stability, height tracking accuracy, and contact behavior.

\subsection{History-Based State Estimation Analysis}
We evaluate the proposed history-based estimator in simulation, where ground-truth base states are available for quantitative comparison.
The estimator predicts the base linear velocity and base height from a short horizon of proprioceptive observations, without direct access to privileged state information.

\begin{figure}[t]
    \vspace{2mm}
    \centering
    \includegraphics[width=1\linewidth]
    {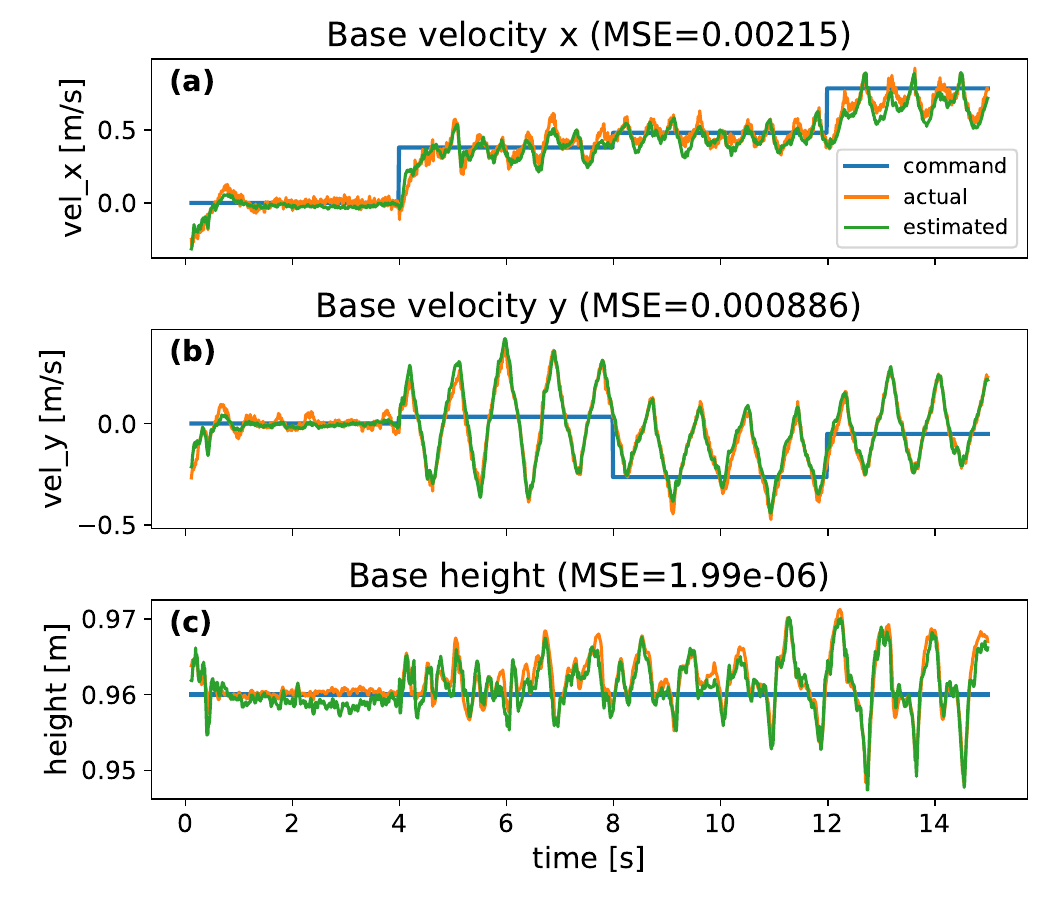}
    \caption{Prediction performance of the history-based state estimator.
Comparison between commanded, ground-truth, and estimated states for (a) base velocity $v_x$, (b) base velocity $v_y$, and (c) base height.
The estimator achieves low mean squared error under dynamic motion conditions.}
    \label{fig:estimation_error}
    \vspace{-2mm}
\end{figure}

Fig.~\ref{fig:estimation_error} presents representative time-series results under dynamically varying velocity commands.
The estimated velocities closely track the ground-truth signals in both magnitude and phase, with no observable drift or delayed convergence during command transitions.
The mean squared errors are $2.15\times10^{-3}$ for $v_x$ and $8.86\times10^{-4}$ for $v_y$, indicating accurate reconstruction of planar motion despite contact-induced disturbances.

For base height estimation, the error remains on the order of $10^{-6}$, demonstrating near-perfect alignment with ground truth.
Such high precision is particularly critical for height-conditioned locomotion, where small vertical deviations may significantly affect contact stability and squatting performance.

Overall, the results confirm that the history-based estimator effectively compensates for partial observability and provides stable, low-variance state feedback for policy execution under dynamic payload conditions.

\begin{figure*}[t]
    \vspace{2mm}
    \centering
    \includegraphics[width=1\linewidth]
    {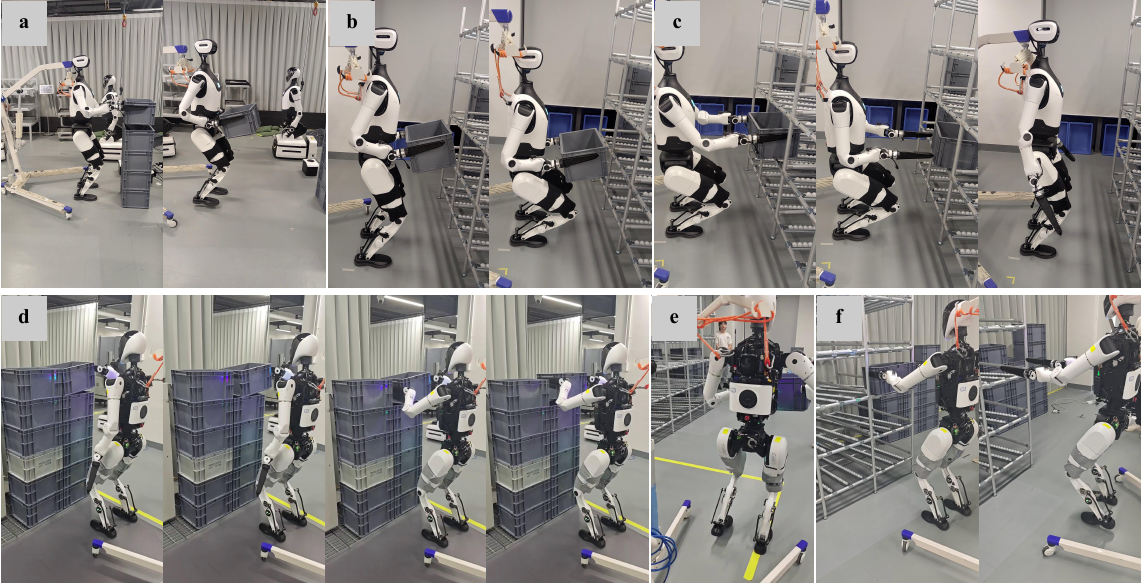}
    \caption{Real-world loco-manipulation experiments in an industrial box transportation scenario. (a--c) Box transportation without depalletizing: the robot transitions from unloaded walking to load-carrying locomotion, squats to a target height, places the box onto a rack while maintaining a crouched posture, and then stands up. (d--f) Transportation with depalletizing: the robot performs box depalletizing, walks while carrying a $5$~kg payload, places the box at the designated location, and continues unloaded walking for the next task. The environment is a one-to-one reconstruction of an industrial factory setting and corresponds to the task scenario of the World Humanoid Robot Games.}

    \label{fig:real_task}
    \vspace{-2mm}
\end{figure*}

\subsection{Real-World Loco-Manipulation Experiments}
\label{subsec:real_world}

We evaluate the proposed framework on real-world industrial box transportation tasks that require tight coordination between locomotion and manipulation.
The experiments are conducted in a one-to-one reconstructed factory environment that replicates an industrial logistics scenario used in the World Humanoid Robot Games.
The robot is required to transport $5$~kg boxes between designated locations, with and without depalletizing, while maintaining stable locomotion under significant load variations and upper-body motion disturbances.

Upper-body manipulation is generated independently using a vision-based pipeline that estimates the 6D pose of the box from a head-mounted depth camera, followed by inverse kinematics for arm control.
The lower-body locomotion policy receives no direct manipulation commands and relies solely on proprioceptive feedback and estimated states to compensate for manipulation-induced disturbances.

Fig.~\ref{fig:real_task} illustrates two representative task sequences.
In the first scenario (Fig.~\ref{fig:real_task}(a--c)), the robot performs a transportation task without depalletizing, transitioning from unloaded walking to load-carrying locomotion, squatting to a commanded height, placing the box onto a rack while maintaining a crouched posture, and then standing up.
In the second scenario (Fig.~\ref{fig:real_task}(d--f)), the robot executes a full depalletizing pipeline, including box removal from a stacked pallet, load-carrying walking, object placement at a target location, and subsequent unloaded locomotion for the next task.

All policies are trained entirely in simulation and deployed on the real robot without additional fine-tuning. Despite the decoupled control architecture, the robot consistently maintains stable walking, smooth foot ground contact, and accurate height regulation across different task phases.
These results demonstrate that the proposed kinematics-based reference with height-conditioned residual control and history-based state estimation enables robust sim-to-real transfer for humanoid loco-manipulation in realistic industrial environments.

\section{Conclusion and Discussion}
\label{sec:conclusion}

This paper presented a load-aware locomotion framework for humanoid robots performing industrial transportation tasks. 
By adopting a decoupled yet tightly coordinated locomotion--manipulation architecture, the approach combines learning-based lower-body locomotion with perception-driven kinematic manipulation, enabling stable walking under varying payloads and upper-body motions.
A kinematics-based locomotion reference conditioned on commanded base height and foot-end trajectories is integrated with residual reinforcement learning control, which accelerates training convergence, improves ground contact behavior, and enhances height tracking compared to purely reward-driven policies. 
In addition, a history-based state estimator augments the policy with inferred base motion states and a compact latent representation, improving robustness to payload variations and manipulation-induced disturbances.

The proposed framework is validated through simulation and real-robot experiments in industrial box-handling scenarios, where policies trained entirely in simulation are deployed without additional fine-tuning. 
Future work will investigate adaptive reference generation and tighter whole-body integration to further improve versatility in complex environments.

\bibliographystyle{Bibliography/IEEEtranTIE}
\bibliography{Bibliography/IEEEabrv,Bibliography/ref}\ 


\begin{IEEEbiography}[{\includegraphics[width=1in,height=1.25in,clip,keepaspectratio]{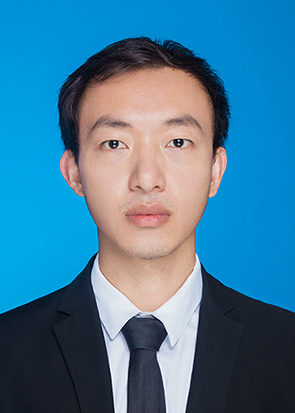}}]
{Lequn Fu}
received the B.E. degree in mechanical engineering from Wuhan University of Technology, Wuhan, China. He is currently pursuing the Ph.D. degree in mechanical engineering at Huazhong University of Science and Technology, Wuhan, China.

His research interests include motion control and learning-based control of legged robots, with particular emphasis on humanoid locomotion, and robust whole-body control for real-world deployment.
\end{IEEEbiography}

\begin{IEEEbiography}[{\includegraphics[width=1in,height=1.25in,clip,keepaspectratio]{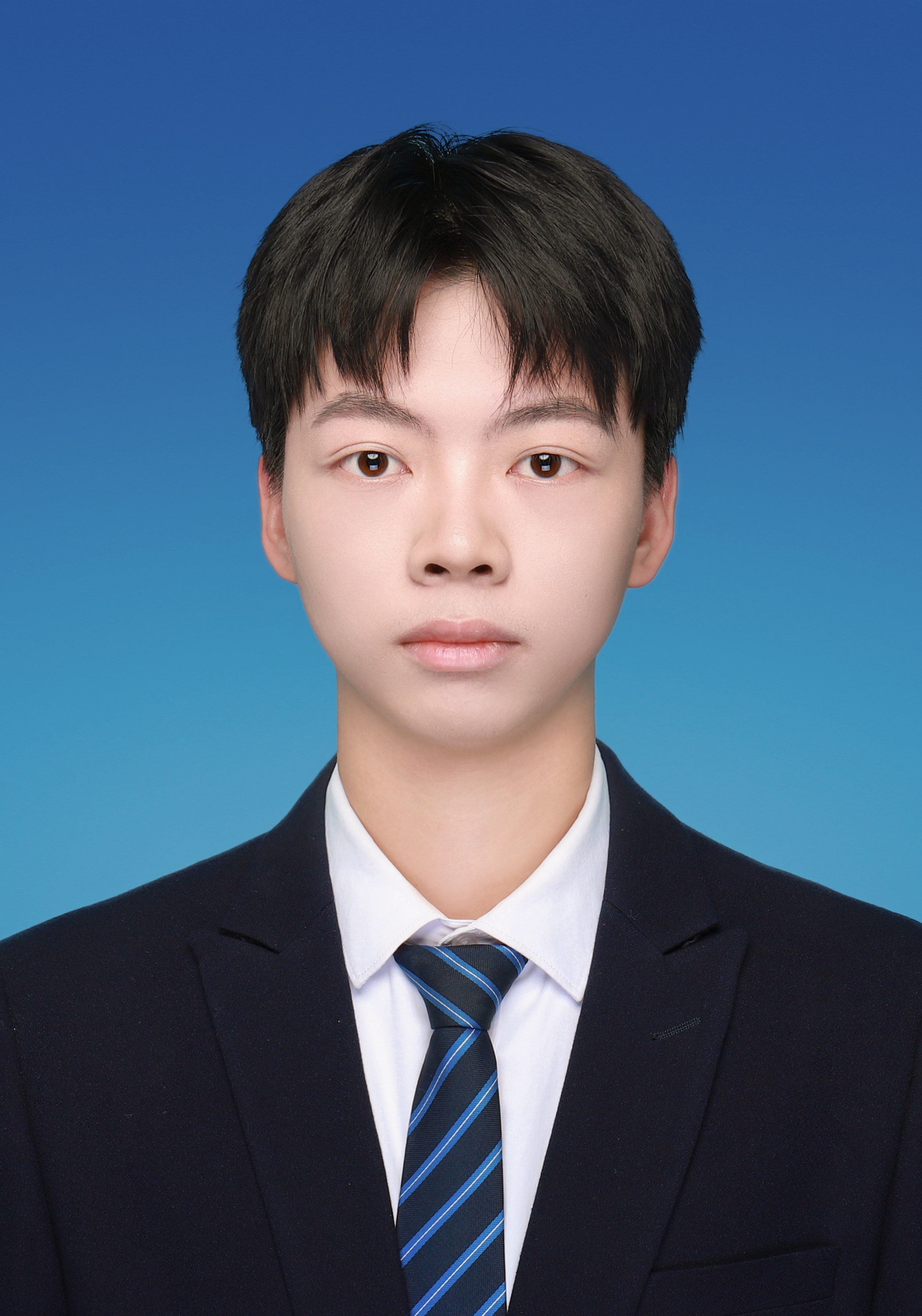}}]
{Yijun Zhong}
received the B.E. degree in industrial engineering from Hunan University, Changsha, China, in 2024. He is currently pursuing the M.E. degree in mechanical engineering at Huazhong University of Science and Technology, Wuhan, China.

His research interests focus on optimization-based motion control for legged robots, particularly motion strategies and gait planning for humanoid robots in complex terrains, as well as control methods to enhance adaptability and stability in unstructured environments.
\end{IEEEbiography}

\begin{IEEEbiography}[{\includegraphics[width=1in,height=1.25in,clip,keepaspectratio]{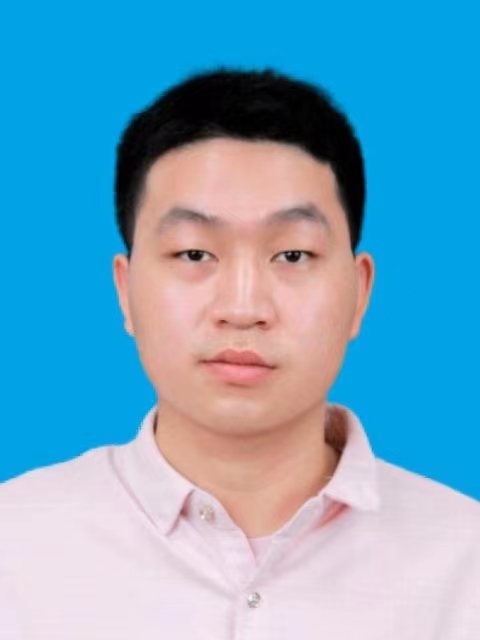}}]
{Xiao Li}
received the M.S. degree from Huazhong University of Science and Technology, Wuhan, China, in 2022. He is currently pursuing the Ph.D. degree in mechanical engineering at Huazhong University of Science and Technology.

His research interests include whole-body coordination control and dexterous manipulation of humanoid robots.
\end{IEEEbiography}

\begin{IEEEbiography}[{\includegraphics[width=1in,height=1.25in,clip,keepaspectratio]{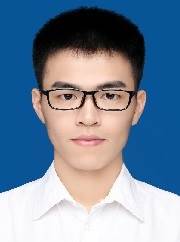}}]
{Yibin Liu}
received the B.E. degree from Hefei University of Technology, Hefei, China, in 2020. He is currently pursuing the Ph.D. degree at Huazhong University of Science and Technology, Wuhan, China.

His research interests include robotic visual SLAM, multi-sensor fusion localization, and reinforcement learning-based visual navigation.
\end{IEEEbiography}

\begin{IEEEbiography}[{\includegraphics[width=1in,height=1.25in,clip,keepaspectratio]{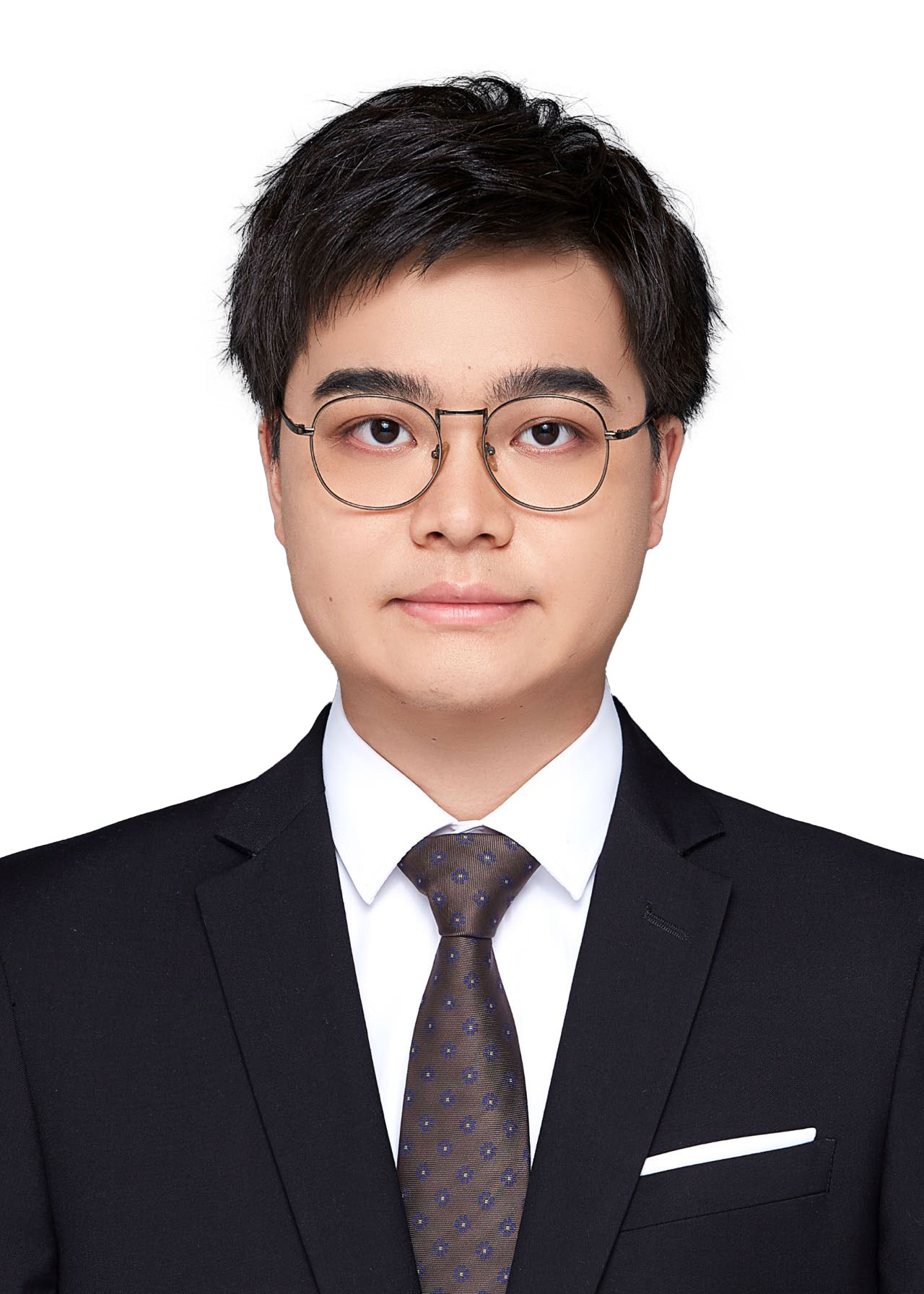}}]
{Zhiyuan Xu}
received his Ph.D. degree in computer information science and engineering from Syracuse
University, Syracuse, NY, USA, in 2021, and his B.E. degree in the School of Computer Science
and Engineering from the University of Electronic Science and Technology of China, Chengdu, China,
in 2015.

He is now with Beijing Innovation Center of Humanoid Robotics. His research interests include Deep Learning, Deep Reinforcement Learning, and Robot Learning, Embodied AI.

\end{IEEEbiography}

\begin{IEEEbiography}[{\includegraphics[width=1in,height=1.25in,clip,keepaspectratio]{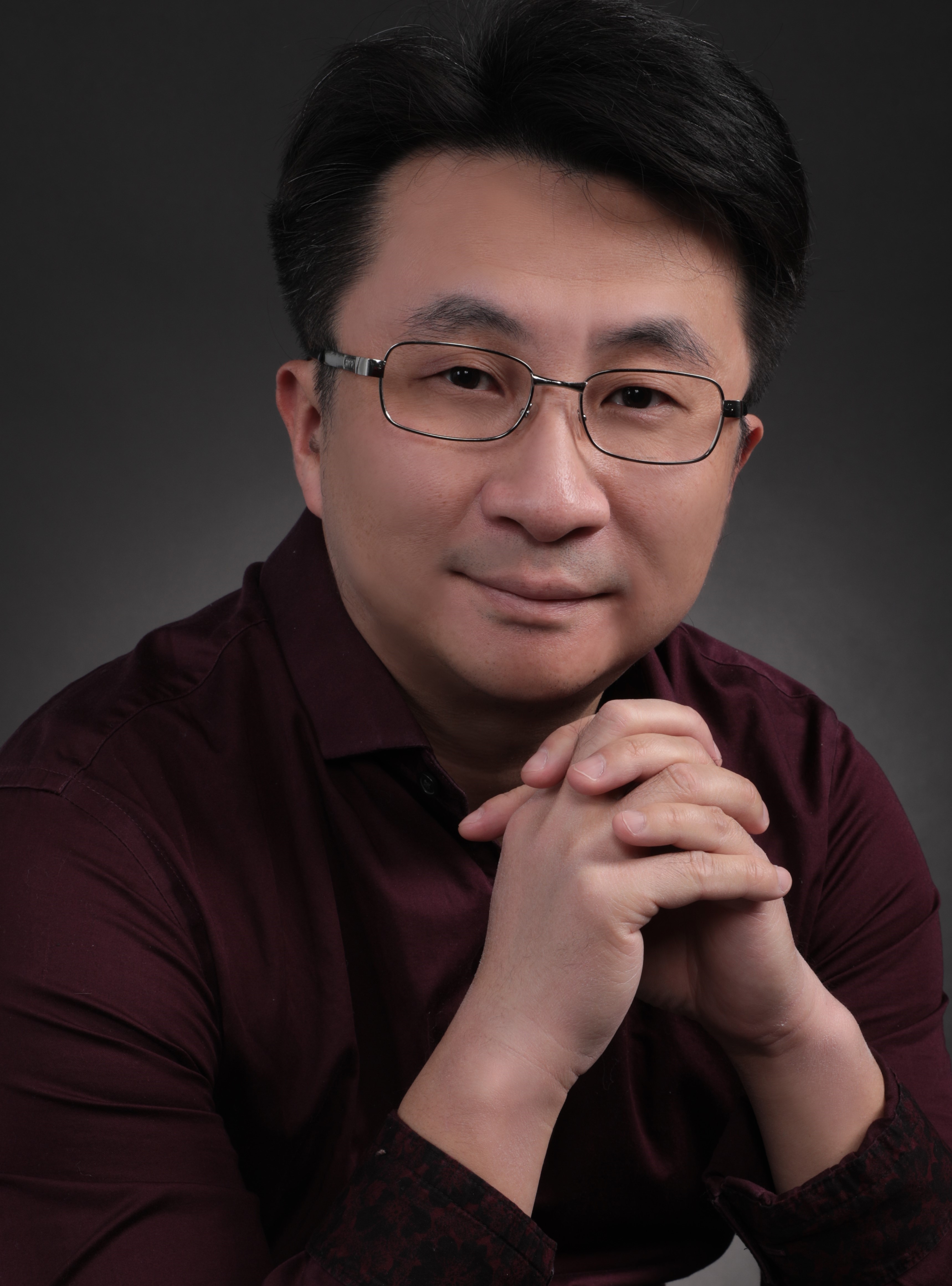}}]
{Jian Tang (Fellow, IEEE)}
received his Ph.D. degree in Computer Science from Arizona State University in 2006. He is an IEEE Fellow and an ACM Distinguished Member. He is with Beijing Innovation Center of Humanoid Robotics.

His research interests lie in the areas of AI, IoT, Wireless
Networking, Mobile Computing and Big Data Systems. He received an NSF CAREER award in 2009. He also received several best
paper awards, including the 2019 William R. Bennett Prize and the 2019 TCBD (Technical Committee on Big Data), Best Journal Paper Award from IEEE Communications Society (ComSoc), the 2016 Best Vehicular Electronics Paper Award from IEEE Vehicular Technology Society (VTS), and Best Paper Awards from the 2014 IEEE International Conference on Communications (ICC) and the 2015 IEEE Global Communications Conference (Globecom) respectively. He has served as an editor for several IEEE journals, including IEEE Transactions on Big Data, IEEE Transactions on Mobile Computing, etc. In addition, he served as a TPC co-chair for a few international conferences, including the IEEE/ACM IWQoS'2019, MobiQuitous'2018, IEEE iThings'2015. etc.; as the TPC vice chair for the INFOCOM'2019; and as an area TPC chair for INFOCOM 2017- 2018. He is also an IEEE VTS Distinguished Lecturer, and the Chair of the Communications Switching and Routing Committee of IEEE ComSoc 2020-2021.
\end{IEEEbiography}

\begin{IEEEbiography}[{\includegraphics[width=1in,height=1.25in,clip,keepaspectratio]{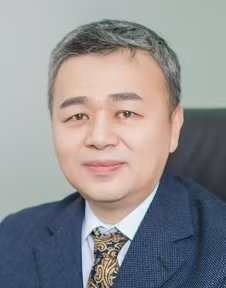}}]
{Shiqi Li}
received the B.E., M.E., and Ph.D. degrees in engineering from Huazhong University of Science and Technology, Wuhan, China, in 1986, 1989, and 1992, respectively.

He is currently a Professor with the School of Mechanical Science and Engineering, Huazhong University of Science and Technology, Wuhan, China. His research interests include humanoid robots and embodied intelligence.

Prof. Li has been actively involved in research and development of advanced robotic systems and intelligent control technologies.
\end{IEEEbiography}

\end{document}